# Assessment of Customer Credit through Combined Clustering of Artificial Neural Networks, Genetics Algorithm and Bayesian Probabilities

Reza Mortezapour
Department of Electronic And Computer
Islamic Azad University
Zanjan, Iran
mortezapour@ymail.com

Mehdi Afzali
Department of Electronic And Computer
Islamic Azad University
Zanjan, Iran
afzali@hacettepe.edu.tr

*Abstract*—**Today, with respect to the increasing growth of demand to get credit from the customers of banks and finance and credit institutions, using an effective and efficient method to decrease the risk of non-repayment of credit given is very necessary. Assessment of customers' credit is one of the most important and the most essential duties of banks and institutions, and if an error occurs in this field, it would leads to the great losses for banks and institutions. Thus, using the predicting computer systems has been significantly progressed in recent decades. The data that are provided to the credit institutions' managers help them to make a straight decision for giving the credit or not-giving it.**
**In this paper, we will assess the customer credit through a combined classification using artificial neural networks, genetics algorithm and Bayesian probabilities simultaneously, and the results obtained from three methods mentioned above would be used to achieve an appropriate and final result. We use the K_folds cross validation test in order to assess the method and finally, we compare the proposed method with the methods such as Clustering-Launched Classification (CLC), Support Vector Machine (SVM) as well as GA+SVM where the genetics algorithm has been used to improve them.**

## I. INTRODUCTION

Today, with respect to the development of database systems and large amount of data saved in these systems, we need an instrument to process the data saved and to provide the users with the information resulted from the process. Data analysis is one of the most important methods that it provides the users and the analysts with some useful models of data with at least intervention of known users in order to make critical and important decisions of organizations according to them. Classification is one of the most common duties of data analysis. In fact, classification has been defined as evaluation of the characteristics of data set and then to allocate them to a set of groups predefined. Data analysis can be used to create a model or a view of a group based on data characteristics by using historical data. Then, we can use the predefined model in order to classify new data sets. Also, we can use it for the future predictions by determining a view that is correspondent with it. Commercial issues such as regression analysis, risk management and case targeting are involved in the classification. In order to overcome the financial problems of credit, organizations and institutions have considered several sections as the credits management. The purpose of the company credits management is to determine policies and to observe strategies that are correspondent with the company's functional aspect in terms of risk and efficiency. If the customers observe the previsions of credit contracts and pay the cash of goods purchased on credit, the company efficiency would be increased. Risk or hazard is a probability that the company credit not be receipted or in order to receipt previous credits, the company would be incurred additional costs.

## II. PREVIOUS RESEARCHES

In the past, many researchers provided traditional statistical methods to credit accounts by using Linear Discriminant Analysis (IDA) and Logistic Regression and it has been used two common statistical methods in the structure of credit rating models. Nevertheless, Krles, Prakash, Reichert and Wagner cho suggested that usually because of considering the classification nature of credit data, IDA is be needed and this fact has been challenged that it seems unlikely to be the covariance matrix of bad and good credit groups.
In addition to IDA method, logistic regression is usually another method for the credit rating. Logistic regression is a model that would be used to predict the probability of an event occurring. This method allows us to use different predictor variables which may be numerical or classified.
Basically, the logistic regression model has initially been used as a method to predict binary outcomes. The logistic regression model doesn't need to the normal multi-variables hypothesis, but it depends on various access of perfect linear relationship between the independent variables for powering the logistic function.

Thomas and West showed that both logistic regression and IDA methods have this tendency to have a linear basic relationship between the variables and thus, it have been reported that it hasn't enough accuracy for rating credit.

Recently, new database search methods can be used to make the credit rating models. Desai, Crook and Over Street used Neural Networks (NW), Logistic Regression and IDA to make the credit rating models. The results indicated that Neural Networks (NN) is showing the expectancy, whereas assessment of bad loans percent performance has been carefully classified. Nevertheless, IRA is as good as NN, whereas the criterion is percent performance of good and bad loans that it has been carefully classified.

West compared the accuracy of credit points of five neural models and he reported that the hybrid structure of neural network models must be considered for the applicants of credit points. In addition, Hu, Kuo and Ho suggested the two-step search method that it uses the self-organizing plan to determine the number of clusters and then the algorithm of K methods would be used to classify the clusters samples. In this study, multiple using of clustering methods and neural networks would be affected by design of credit points' model.

Malhotra and Malhotra compared the operation of Artificial Neural-Fuzzy Interface System (ANFIS) and different models of Discriminant Analysis to potential defaults screening of customer's loans. This result reported that in order to identify the bad credits demand, ANFIS is better than different methods of discriminant analysis.

In recent years, the Support Vector Machine (SVM) was introduced to investigate the problems of classification (the demand for a new classification method). Many researchers used SVM method to rate the credit and to predict the financial risks, and the results obtained were promising. In addition, Hung, Chen and Wang chose three strategies to make the hybrid models of SVM-based credit points and to investigate the customer's credits points through the characteristics of customer input.

## III. THE PROPOSED METHOD

In this study, a method has been proposed to assess the customer's credit that it uses three classifiers including Artificial Neural Networks, Genetics Algorithm and Bayesian Classifier, and then it extracts the final result obtained from above methods by a mechanism.

Fig. (1) shows the workflow of the proposed method. In the following sections, we will describe each section. Of course, due to the clearness of Bayesian classifier, we will not describe this issue and will express the experiments and the results at the end of this paper.

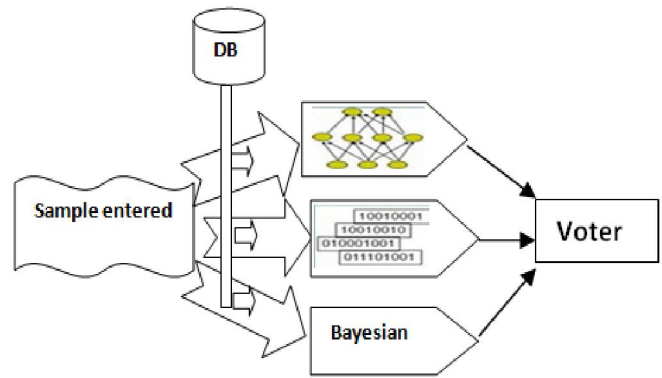

Figure.1 Workflow of the proposed method.

### A. Artificial Neural Network

After back-propagation training, multi-layer perceptron networks are usually considered as a sample of standard networks for modeling the prediction and the classification: selecting an optimal MLP architecture is one of the areas that has been studied.

Method of function of multi-layer perceptron neural network with back-propagation training essentially consists of two main paths. The first path is called forward path where the input vector is applied to MLP network and its effects would be propagated from the middle layers to the output layers. The output vector formed in the output layer is true response of MLP network. In this path, the network parameters would be considered constant and invariable. The second path is called backward path. Unlike the forward path, the parameters of MLP network would be changed and adjusted in this path. This adjustment would be done according to the error correction code. The error signal is formed in the output layer of the network. The error vector is defined as the difference between the optimal and the true response of the network. After calculating in the backward path of output layer and through the networks layers, the amount of error would be distributed in the entire network. Since the recent distribution has been done in a path contrary to the weight communications of synapses, the term **back error propagation** has been selected in order to describe the behavioral modification of the network. The network parameters would be adjusted in such a way that the true response of the network is as more optimal as possible. After making a multi-layer perceptron neural network through the back-propagation training, some decisions must be made that they have been shown in the following.

#### 1) Activation Function of Neurons

In a typical application that several inputs have been coded as 0-1, the neuron outputs are 0-1 with the annular activation functions and they are approximately -1 and +1 with the hyperbolic tangent activation function. In this condition, the hyperbolic tangent is the best option. We used the annular and the hyperbolic functions in German and Australian datasets, respectively.

### 2) Learning Rules

The main rule of learning that has been provided by Rommel Hart is called Delta Rule. This rule is a way that the artificial neural network learns from his mistakes, and common processing for this learning consists of three actions.

a) Outputs Calculations

b) Comparing Outputs with Optimal Solutions

c) Adjusting the Weights and Bias Values and Repeating the Processing

Learning usually begins by randomizing the weights and the bias values. The difference between the real output and the optimal output is called delta. It is better to minimize the delta. Decreasing the delta would be done through the weights and bias values, which this kind of learning is so-called Supervised Learning too. In this kind of learning, the important and the effective issue in the final results is the number of repetitions. Table (1) shows different repetitions of the network and validities obtained to illustrate this issue. There is another kind of learning that is called Unsupervised Learning. In this kind of learning, just the network input would be motivated and the network is self-organizing. In this way, the network would be structuralized internally. Thus, each hidden processing element responses strategically to a different set of input stimuli and it is not be used any knowledge to classify the outputs. In this way, there is this probability that the network can or can't produce a meaningful issue to everyone who is learning. Some examples of this learning are Correspondence Theory and Kohonen's Self-organizing Maps.

| Number of occurrences | Australian | German |
|---|---|---|
| 300 | 75.07 | 73.4 |
| 500 | 67.54 | 73.4 |
| 1000 | 70.14 | 73.4 |
| 2000 | 74.93 | 78.6 |
| 2500 | 71.01 | 78.2 |
| 3000 | 75.65 | 81.4 |
| 4000 | 78.12 | 80.2 |

Table 1: Number of iterations in training and the accuracy obtained.

### 3) Learning Rate

Learning rate is the last key that must be determined for decision-making. Most of the people used the learning rate in a way that they choose it with a large number close to 1. Optimal learning rates resulted from the smooth level of RMS error. If the graph of RMS error has the high increasing and decreasing variations in the output layer, it is clear that the learning rate utilized is not optimal and it should be decreased equally to all of the layers. We have set the learning rate for both datasets equal to 0.7.

### B. Genetics Algorithm

#### 1) Determining The From Of Solution In The Genetics Algorithm

In the standard form of the genetics algorithm, the solutions are as the binary strings, but using this form for many problems leads to complicate the solutions and in many cases, providing the solution in this form will be impossible. Therefore, in the genetics algorithm applications for the optimization problems, instead of using the complicated binary strings, we used a solution form corresponded with the proposed problem. In this problem, we also used the solution form of the problem.

#### 2) Method Of Determining Initial Population Of The Genetics Algorithm

In the standard genetics algorithm, the initial population would be achieved randomly. This method may be appropriate for unlimited problems. But in some other problems, the initial solutions can't be determined randomly, because there is no guarantee to exist the solutions. Thus, we have to select the initial population in a way that all of the solutions are justified. We also used the other methods' training data for the initial population.

#### 3) Genetic Operations

We usually try to choose the operations in a way that the proportion rate of new responses (Children) is better than the parents. In the genetics algorithm utilized, we used two-point mixture operator as well as the mutation operator in the fields that were possible.

#### 4) Recognition and Selection

So far, we have achieved three groups of responses by adjusting the parameters required for three methods used in the proposed method. In each method, graphs 1 and 2 of the validity obtained have been represented for German and Australian datasets. Now we extract the response. We test three methods for the final result. The first method is to use the majority voting, the second is to use a neural network and the third method is to use weighting for each of the methods. In this method, if we consider each of the methods as "V", we can use the following formula to extract the final result:

$$R = \sum_{i=1}^{n} v_i \qquad (1)$$

where R is the result, n is the number of methods and V is the result of method obtained. Table 2 shows the results obtained from three methods mentioned above for two datasets.

## IV. EXPERIMENTS

All of the results provided are resulted from running the programs on a system having characteristics such as Memory 3GB, Intel Pentium 2.2 GHZ and XP operating system. We used MATLAB and VB.Net 2008 programming language to implement the program. In order to certify this methods, we used the k_fold cross validation in the results provided, where k is equal to 10.

For all the tests, we used two datasets that their characteristics have been represented in Table 2.

| Dataset Name | Record number | number of numeric field | number of non-numeric field | Record number of class 1 | Record number of class 2 |
|---|---|---|---|---|---|
| Australian Credit Approval | 690 | 6 | 8 | 307 | 383 |
| German credit dataset | 1000 | 7 | 13 | 700 | 300 |
| German credit dataset-numeric | 1000 | 24 | 0 | 700 | 300 |

Table. 2 Characteristics of data sets used in the paper

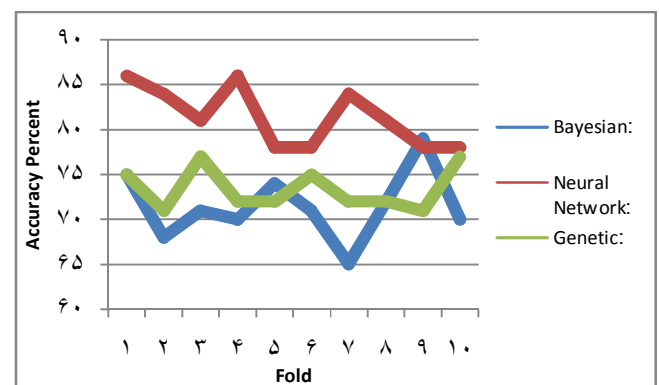

Figure. 1 Results of different methods on the German dataset

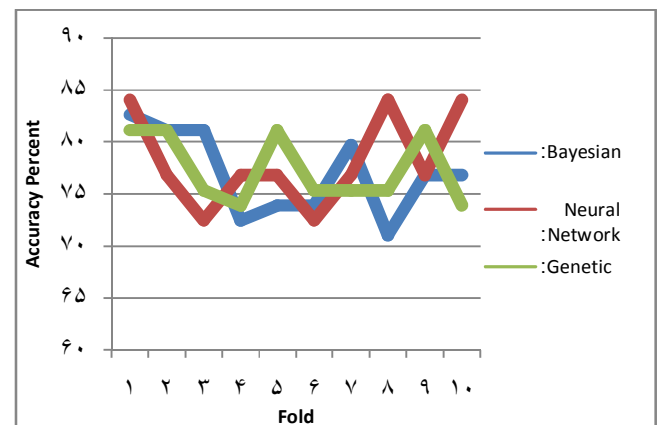

Figure. 2 Results of different methods on the Australian dataset

Since each algorithm had a different result compared to other algorithms, thus we have linked each of the algorithms' output to a voting system that has been shown in Fig. (1). Here, we considered three strategies to the voting system that Table 2 shows the results obtained from each of the strategies and their comparison to the previous three methods. We worked in a method that in the first method, the majority voting was done; that is each of the algorithms has the same impact on the output. In the second method, the algorithms' outputs would be entered in a multi-layer perceptron neural network and with respect to the learning rate of 0.7 that is considered to the network, it represents the results of the output mentioned above, which it has an appreciable improvement compared with the majority voting. The third strategy is based on an appropriate weighting to each algorithm; that is with respect to the results obtained from each algorithm and its impact on the final result, we choose a weight according to it. In this method, we have set the weight of 0.5, 0.29 and 0.21 to neural network, Bayesian and genetics algorithm method, respectively.

## V. RESULTS

In this system, regarding the turnovers conducted and their impact on the refund of the credit allocated to previous customers or all of the people that their information is available, we have evaluated the importance of each item and thus, we have omitted incorrect relationships and characteristics. In this paper, we have provided a comprehensive system to assess the customer's credit that it can significantly solves the problems of existing systems. This system can assess the credit and it can appropriately distinguish the credit decision-makings with a high accuracy despite registering incorrect information in data entry due to using different techniques and methods of data analysis. The proposed system has not environmental dependency; that is we can use this system in any environment due to the need for primary data. This system can provide different assessments for political and military applications in order to find the credit of proposed sections according to the activity has been asked.

| The method used | Australian | German |
|---|---|---|
| Majority voting method | ٧٨٫٢ | ٨٣٫٣ |
| Artificial neural network method | ٨٨٫٩٥ | ٨٧ |
| based on the weighted voting method | ٨٤٫٧ | ٩٠ |
| Clc | 86.52 | 84.80 |
| Mysvm | 80.43 | 73.70 |
| GA+svm | 86.90 | 77.92 |

Table. 2 Comparison of Methods